\documentclass[preprint,12pt]{elsarticle}

\usepackage[T1]{fontenc}
\usepackage[utf8]{inputenc}

\usepackage{amsmath,amssymb,amsfonts}
\usepackage{algorithmic}
\usepackage{graphicx}
\usepackage{textcomp}
\usepackage{tabularx}
\usepackage{tabulary}
\usepackage{multirow}
\usepackage{caption}
\usepackage[hyphens]{url}
\usepackage{subcaption}
\usepackage{textcomp}
\usepackage{wrapfig}
\usepackage{tikz}

\usepackage{xspace}

\graphicspath{{img/}}

\journal{Engineering Applications of Artificial Intelligence}

\newcommand{\shadowwolf}{\textit{ShadowWolf}\xspace}
\newcommand{\mainzaun}{\textit{mAInZaun}\xspace}
\newcommand{\fone}{\(\text{F}_1\)\xspace}
\newcommand{\wolfornot}{\textit{Wolf-or-Not}\xspace}

\newcommand*\circled[1]{\tikz[baseline=(char.base)]{
            \node[shape=circle,draw,inner sep=0.5pt] (char) {#1};}}

\begin{document}
\begin{frontmatter}
\title{ShadowWolf -- Automatic Labelling, Evaluation and Model Training Optimised for Camera Trap Wildlife Images}

    \author[label1]{Jens Dede}
    \ead{jd@comnets.uni-bremen.de}
    \author[label1]{Anna Förster}

    \affiliation[label1]{organization={Department of Sustainable Communication Networks, University of Bremen},
    addressline={Bibliothekstr.\ 1},
    postcode={28359},
    state={Bremen},
    city={Bremen},
    country={Germany}
    }


\begin{abstract}
    The continuous growth of the global human population is leading to the expansion of human habitats, resulting in decreasing wildlife spaces and increasing human-wildlife interactions. These interactions can range from minor disturbances, such as raccoons in urban waste bins, to more severe consequences, including species extinction. As a result, the monitoring of wildlife is gaining significance in various contexts. Artificial intelligence (AI) offers a solution by automating the recognition of animals in images and videos, thereby reducing the manual effort required for wildlife monitoring. Traditional AI training involves three main stages: image collection, labelling, and model training. However, the variability, for example, in the landscape (e.g., mountains, open fields, forests), weather (e.g., rain, fog, sunshine), lighting (e.g., day, night), and camera-animal distances presents significant challenges to model robustness and adaptability in real-world scenarios.
    
    In this work, we propose a unified framework, called \shadowwolf, designed to address these challenges by integrating and optimizing the stages of AI model training and evaluation. The proposed framework enables dynamic model retraining to adjust to changes in environmental conditions and application requirements, thereby reducing labelling efforts and allowing for on-site model adaptation. This adaptive and unified approach enhances the accuracy and efficiency of wildlife monitoring systems, promoting more effective and scalable conservation efforts.
\end{abstract}


\end{frontmatter}



\section{Introduction}
\label{sec:introduction}

As human populations continue to grow \cite{un2024worldpopulation}, the number of natural habitats is getting smaller, reducing the space available for wildlife. This often results in negative consequences such as species extinction and human-wildlife conflicts, like elephant-human conflicts in Sri Lanka and India \cite{10.3389/fevo.2018.00235} or wolf and bear  conflicts in Europe \cite{fernandez2016conflict}. The conflicts are mostly of economic nature, where animals destroy agricultural fields, attack livestock and domestic animals, or destroy infrastructure such as streets, power lines, etc. Many mitigation techniques are known, from economic compensation through resettlement to repellent actions. Each of them have their advantages and disadvantages, with repelling actions being the most promising in terms of preserving nature balance and supporting a human-wildlife coexistence. However, in order to efficiently repel the problematic animals, an automated animal detection is desirable, so that reppeling is conducted in a specific, effective and safe way.

One effective approach for wildlife monitoring and species identification is the use of camera traps. These devices activate automatically, capturing images based on specific settings, and can quickly generate a large amount of images. In our \mainzaun project\footnote{\url{https://intelligenter-herdenschutz.de/}}, we employ a specialised version of these traps to detect wolves in their natural habitats using artificial intelligence (AI). Our target is to detect the wolves around livestock fences in the wild and automatically start repelling actions. In this paper, we focus on the robust detection in various environmental conditions. 

Animal detection has been well explored in recent years, especially with the rise of Deep Learning Artificial Intelligence (AI) techniques\cite{rigoudy2023deepfaune,norouzzadeh2018automatically,beery2019efficient}. 
Training such an AI model typically involves three steps: collecting example data, labeling the class of interest (e.g., "wolf"), and then training a detection model.
However, outdoor images show unique challenges. For example, sunny days in a forest during the summer and dusty days in the autumn in a mountainous area result into completely different images even when showing the same animal. Additionally, animals of interest are not always the central focus of the image and may even be partially obscured, further complicating detection. Especially for our purposes to protect livestock in various different environments, this straight-forward approach does not work well. Our envisioned tool needs to work with various backgrounds, in different seasons, different appearances of the animals (winter-summer fur, fur colouring, etc.), different lighting conditions, etc. It is merely not possible to gather training data for all of these cases, especially as wild wolves are not very cooperative in posing for images. 

Another challenge is the fast progress of AI techniques, offering new models and tools at very high turn-over rates. For practitioners, this means a very tedious and manual process of re-building the models and the pre- and post-processing steps required. 

To address these challenges, we propose our unified framework called \shadowwolf\footnote{\url{https://github.com/ComNets-Bremen/ShadowWolf}}. This framework tackles those key issues by the following measures:

\begin{itemize}
    \item \textbf{Incremental Retraining:} Offers quick adaptation to new environments and species.

    \item \textbf{Modular Approach:} Allows quick implementation and testing of new ideas, models and approaches for the image processing. It also enables easy comparison of new steps with existing ones based on end-to-end performance evaluation.

    \item \textbf{End-to-End Performance Evaluation:} Facilitates evaluation of the impact of new algorithms on the complete image recognition chain, ensuring overall effectiveness including the pre- and postprocessing.

    \item \textbf{Autonomy:} The toolchain is almost entirely autonomous, making it practical to adapt to new environments and add new object classes with minimal human intervention.

    \item \textbf{User-Centered Crowd-Sourcing:} The only non-autonomous step involves user-centered crowd-sourcing, allowing for rapid evaluation of large new datasets and image recognition results in short time.
\end{itemize}

This toolchain not only solves our problem of wolf detection in the wild with camera trap images, but has the potential to increase and ease the applicability of automatic image detection in many other areas, such as smart city traffic detection, smart farming, etc. It enables practitioners to continuously refine their models, integrate new images seamlessly and automate most of the necessary steps. 

The remainder of this paper is structured as follows: This section defines the problem and lists the current state of the art. It also details the objectives of this work. Section~\ref{sec:methodology} gives more background: How did we collect our data, which models exist, and what metrics we use? We introduce the system architecture of \shadowwolf in Section~\ref{sec:architecture} and evaluate its performance in Section~\ref{sec:performance}. We discuss the results in Section~\ref{sec:discussion}, and conclude this work in Section~\ref{sec:conclusion}. In the appendix, we offer additional information and data to our work.

\subsection{Problem Statement}
\label{sec:problem_statement}

Working with camera trap images using AI to detect animals presents several challenges. First, the quality of the images highly depends on environmental conditions such as variations in weather (rain, snow, fog) or changing light conditions (daylight, nighttime, overcast skies). This can make images of the same species appear vastly different as discussed in our previous work \cite{dede2023aiinthewild}. Second, these images vary greatly depending on the environment or landscape (mountains, forests, flatlands), background, and the specific animals being detected. Third, adapting a detection model to a new species or camera position often requires the interaction of a domain expert to mark and identify the species and create labelled training data. Lastly, the animals of interest are not always the dominant part of the image or sometimes even partly hidden.

In real-world scenarios, a single-shot training to recognize wolf images is not practical. In our case, for example, we envision a product to protect livestock from wolves in various natural environments. This product needs to be pre-trained for some known environments, so that it can start working straight out of the box and with a very short installation duration. While already working, the system needs to gather additional images, continuously evaluate its own precision and accuracy and adapt accordingly. While ongoing research focuses on making the ML models more and more precise and to train them faster and more energy-efficient, they do not solve our problem of continuous online adaptation. 

In this work, we address all these challenges with our \shadowwolf, which offers a framework to quickly adapt object detection models for wildlife monitoring, thereby reducing the workload for domain experts. The foci are on reduced effort for the labeling and an accelerated model adaptation to new environments or application scenarios.

\subsection{State of the Art}
\label{sec:sota}

In this paper, we focus on the recognition of animals, particularly wolves, in camera trap images captured in the wild. Three key areas are central to our research and discussed in this section. First, we examine prior efforts in recognizing animals in the wild from images. Second, we highlight the importance of training and continuous adaptation to accommodate new species and environments. Third, we review the tools and algorithms available for training image recognition models, including support tools like labeling and pre-processing. This section offers an overview of these fields to contextualize our study, though it is not intended to be exhaustive.

\subsubsection{Animal Recognition from Images}

The authors in \cite{norouzzadeh2018automatically} introduce a project to count and identify wild animals in camera trap images, focusing on 48 species from the Snapshot Serengeti dataset \cite{serengeti_dataset}. They use a two-stage approach: first, their model detects whether the picture shows an animal or is empty; second, it identifies the species in the image. From this information, they also count the animals and add attributes like \textit{standing} and \textit{eating}. They implemented six deep learning architectures using TensorFlow and compared their performance on the dataset, achieving over 90\% accuracy in animal identification. However, their approach lacks a mechanism for incorporating new object classes or data to adapt to evolving environments (unlike the relatively stable climate and background of the Serengeti) which is the focus of our work.

The same authors proposed an active learning approach for camera trap images in \cite{norouzzadeh2021deep}, employing a three-step pipeline and a human oracle to guide the process. They utilize a Faster-RCNN (ResNet-50) model, implemented in PyTorch, for detection. Their primary goal was to minimize human interaction time by preselecting relevant images rather than adapting the detection model. Using this active learning strategy, they successfully reduced the annotation bottleneck by 99.5\% for their datasets, significantly lowering the workload for human expert annotators.

In contrast, our approach leverages a crowd-based system where real humans, who are not necessarily experts, evaluate detections to improve overall accuracy. This method broadens accessibility and enhances detection performance through collective input.

In \cite{tabak2019machine}, the authors trained a classifier using the ResNet-18 architecture in R. Using camera trap images from North America, they achieved an accuracy of 97\%. Unlike object detection algorithms, their approach does not return bounding boxes, i.e., positions of animals in the images, but provides a general overview of what is visible. Our work focuses on identifying individual animals rather than providing a summary of the whole image.

The work in \cite{beery2019efficient} focuses on recognizing any kind of animal in new environments. They create boxes of interest without identifying specific objects or class types. This output helps analyze images generally and could serve as a preprocessing step for our \shadowwolf.

As can be seen, many researchers have demonstrated the potential of image recognition for animal identification and counting. While the models perform well, adding new object classes or adapting to new environments, such as different seasons or habitats, remains challenging. Another significant issue is labeling images for training, which is crucial for good performance. Labeling is still a tedious, manual task that is expensive and does not scale well.

\subsubsection{Active Learning}
(Semi-)automatic adaptation of object detection models to new classes or environments is a prevalent challenge, particularly in domains like self-driving cars. For instance, the authors in \cite{9304565} employ active learning to retrain object detection networks for self-driving applications. They introduce an uncertainty index and evaluate two active learning methods that prioritize hard-to-detect classes. However, their work does not address the challenge of detecting hard-to-see objects or provide an option for incorporating human verification into the loop, which are central aspects of our approach.

In \cite{s24051509}, the authors propose a human-in-the-loop framework to enhance object detection in remote sensing (e.g., satellite imagery). Their approach involves domain experts labeling images, with crowdsourcing used to assess the quality of these annotations. This framework improves object detection accuracy and enhances the generalization of the models. Our work differs in several key aspects. First, we focus on camera trap images, which are fundamentally different from satellite imagery. Second, we address the unique challenges posed by camera trap data by introducing preprocessing steps to standardize the highly diverse images. Finally, instead of relying on domain experts, we integrate crowdsourcing with detection outputs to refine or dismiss results, thereby eliminating the need for domain expertise.

Active learning, along with dynamic adaptation of training data and human-in-the-loop technologies, remains an area of active research. To the best of our knowledge at the time of writing, no existing work focuses on our type of images, removes the reliance on domain experts, and offers a fully dynamic retraining pipeline.

\subsubsection{Tools for Image Recognition}

Machine learning, especially deep learning with neural networks, has been known for several decades. In the past two decades, the required computing power has been met with the rise of GPUs, combined with NVIDIA's CUDA, and Google's Tensor Processing Units (TPUs). These hardware accelerators are optimized for training and running complex neural networks significantly faster than regular CPUs.

Image processing, specifically automatic classification of images and detection of objects in pictures and video sequences, is advantageous for many applications. Frameworks like Google's TensorFlow \cite{tensorflow2015-whitepaper} and PyTorch \cite{NEURIPS2019_9015} by META/Facebook help build, evaluate, and deploy new algorithms, making them accessible even to non-experts in the field of computer vision.
Using these frameworks, one can quickly analyze thousands of photos for the presence of specific objects or classes, which is ideal for environmental monitoring, such as using wildlife cameras.

Tools like CVAT \cite{CVAT_ai_Corporation_Computer_Vision_Annotation_2022}, LabelImg \cite{tzutalin2015labelimg}, and Timeworx\footnote{\url{https://timeworx.io/}} assist in annotating images. They support humans in labeling images or delegate the work to others. Some tools can also use models to detect objects for annotation, further reducing the workload for annotators. However, these tools still require a comparatively high effort from users, such as drawing boxes and checking results in specialized software.

In contrast, our approach minimizes this effort by leveraging crowdsourcing: Using our \wolfornot, anyone can vote on what they see in a particular image. We use a majority vote to detect this and combine this user knowledge with machine learning findings.
To the best of our knowledge, no other toolchain combines arbitrary machine learning models with crowd-sourced user validation to automatically generate training data and iteratively and automatically create new models.

\subsection{Objectives of this Work}
\label{sec:objectives}
\begin{figure}
    \centering
      \includegraphics[width=0.75\columnwidth]{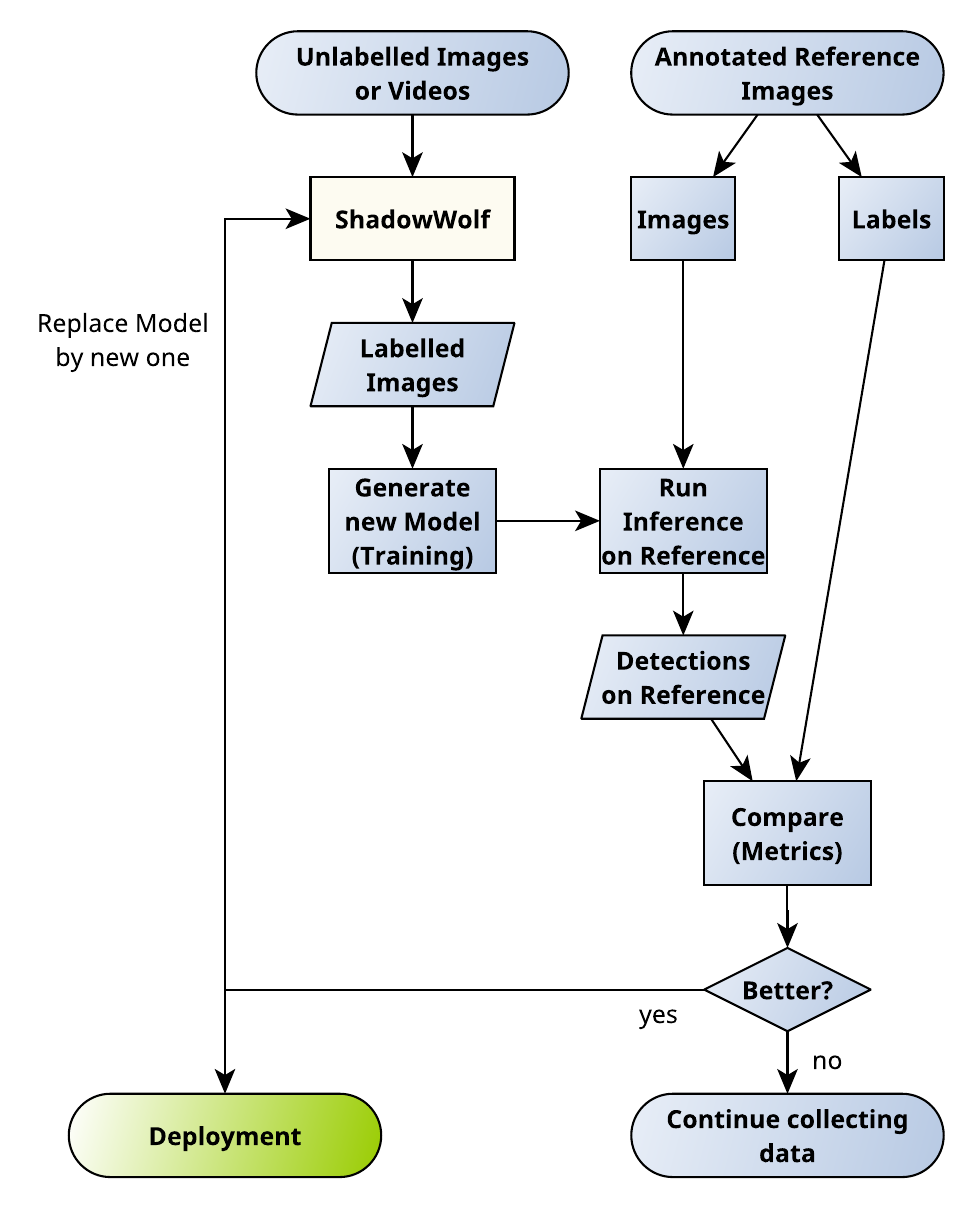}
      \caption{Our iterative approach utilizes the \shadowwolf model to label images. The generated labels are then used to train a new model generation. Subsequently, we evaluate the new model's performance against a reference dataset. If the performance shows improvement, the model is considered better. If not, we continue collecting additional data and repeat the process. In case of an improved model, we use this in \shadowwolf and can also deploy it to other systems like detection in the field as marked by the green box.}
      \label{fig:cont_learning}
\end{figure}

Gaining a deeper understanding of wildlife is essential to foster the coexistence between humans and animals. Camera traps are an effective way to observe animal behavior, track their presence, and map their movement paths. However, the vast volume of images captured by these devices can overwhelm human analysts. With the development of \shadowwolf, we can reduce the workload on domain experts by minimizing the amount of manual labeling required. We create an efficient process to adapt the detection models to new environments. This iterative process, illustrated in Figure~\ref{fig:cont_learning}, is central to the effectiveness\cite{dede23-animals-wild}. By using \shadowwolf to detect animals and generate annotations iteratively, we establish a continuous learning cycle that allows models to adjust to specific wildlife behaviors and habitats over time.

After each successful round, i.e., processing a batch of images leading to a better performing model, we have two outputs:

\begin{enumerate}
    \item A model performing better on the reference dataset.
    \item The training data used to train this model.
\end{enumerate}

Especially the latter can be used to train arbitrary models suitable for various application scenarios ranging from embedded systems running minimalistic models for battery powered in-field systems to fast, GPU-powered systems for large data analysis. This step is indicated by the green deployment box in Figure~\ref{fig:cont_learning}.

Through the reduction of manual labeling effort and the streamlined model adaptation workflow, \shadowwolf makes it feasible to analyze large datasets with minimal human interaction, giving a better understanding of animal behavior and supporting conservation efforts. Ultimately, this supports anyone working with camera-trap images of wildlife and helps bridge the gap between technology and nature to enhance coexistence between human and animal populations.

\section{Methodology}
\label{sec:methodology}
The \shadowwolf model, as introduced in this work, requires input images from camera traps and an object detection model to identify occurrences of animals. To evaluate the output, which consists of annotated images, we employ specific metrics, all of which are introduced in this section.

\subsection{Data Collection}
\label{sec:datacollection}

The realistic implementation and evaluation of our \shadowwolf requires a certain amount of images. The objective is to work with realistic camera trap images, which come with various drawbacks, rather than perfect, high-quality images -- of which several datasets already exist. Since most publicly available datasets do not include these less-than-ideal images, we decided to collect our own dataset. We have previously discussed the data collection process and its challenges in detail in \cite{dede23-animals-wild,dede2023aiinthewild}; here, we provide a brief recap of the technology used and the characterization of the dataset.

For the data collection, we used an off-the-shelf surveillance camera, the Axis M2025-LE, equipped with infrared capabilities for night operation and full HD resolution. This camera was placed in nearby zoos within wolf enclosures, capturing two images per second when motion was detected. Recording began one second before motion was detected and ended five seconds after motion ended. Additionally, one image per hour was taken as a reference and for system monitoring. To reduce false detections, areas not of interest such as treetops were excluded.

Despite the increasing number of wolves in Germany, capturing a sufficient number of images in the wild remains challenging. Therefore, we collaborated with wildlife parks hosting wolves and set up our camera systems there. The parks, dates of data collection, and the number and size of images are detailed in Table~\ref{tab:parks}. This study focuses on two parks -- \textit{Wingster Waldzoo} and \textit{Wildpark Lüneburger Heide} -- both home to European Grey Wolves, similar to those found in the wild in Germany. The cameras were mounted on the fences surrounding the wolf enclosures.

\begin{table}
    \centering
    \begin{tabularx}{\columnwidth}{ m{7.5em} l l l X }
        &                    &         \multicolumn{2}{c}{\textbf{Images}} &\\
        \textbf{Park} & \textbf{Timespan} & \textbf{Number} & \textbf{Size}\\
    \hline
        Wildpark Lüneburger Heide& 2022-03-28 -- 2022-03-30 & 27543&47.5GB\\
        \hline
        Wingster Waldzoo&2023-01-11 -- 2023-02-02 & 63249&91.5GB\\
    \hline
\end{tabularx}
    \caption{The parks where we collected our dataset.}
    \label{tab:parks}
\end{table}

\begin{figure}
  \centering
    \includegraphics[width=1\columnwidth]{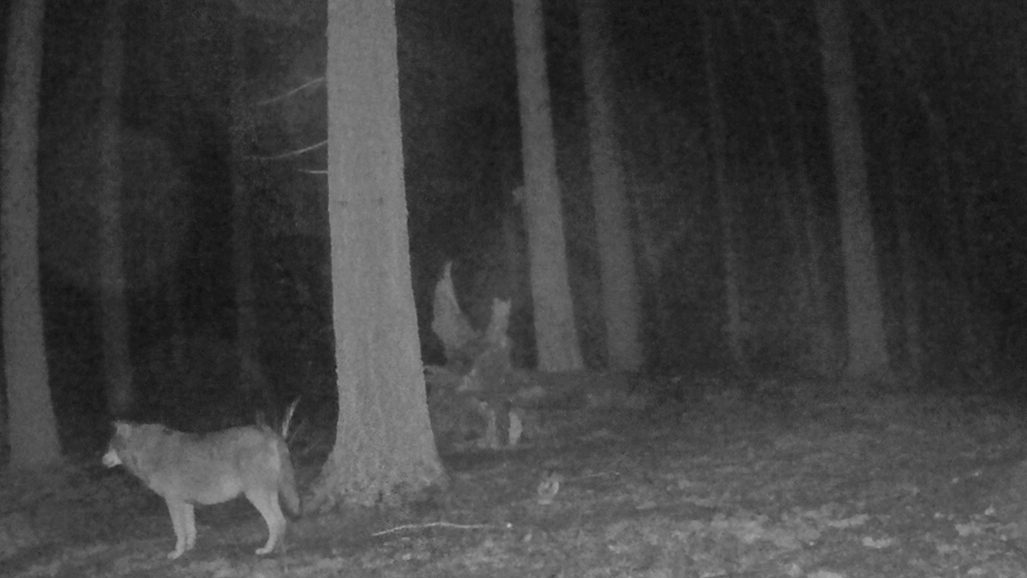}
    \caption{A wolf captured in the night. This picture was taken in the \textit{Wingster Waldzoo}.}
    \label{fig:night_wolf}
\end{figure}

\begin{figure}
  \centering
    \includegraphics[width=1\columnwidth]{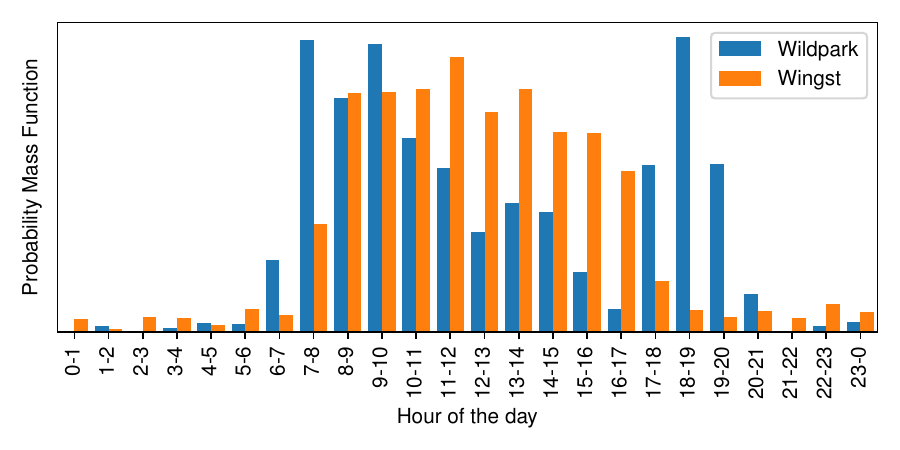}
    \caption{The distribution of our image dataset over a day varies between the parks. During the night, images are only recorded when motion is detected within the range of the built-in infrared spot.}
    \label{fig:time_distribution}
\end{figure}

Few images (less than 0.1\%) were corrupted and subsequently removed. Finally, our dataset consists of  90,750 images, with 79,638 (88\%) taken during the daytime and 11,112 (12\%) at night in infrared mode, for example, as illustrated in Figure~\ref{fig:night_wolf}. The number of images captured is influenced by wolf activity and camera range, which can be affected by environmental conditions like fog, rain, and snow. Figure~\ref{fig:time_distribution} shows the distribution of images captured at different times of the day for each park.

The \textit{Wildpark} data reveals two peaks in activity: one in the morning and another in the afternoon/evening. Our largest dataset, from \textit{Wingst}, shows activity during main opening and working hours from 07:00 to 18:00.

From these images, we create our test dataset, that includes images from various angles and under different lighting and weather conditions, while maintaining the time series. For the Wildpark dataset, one time series per hour was selected, approximately 30 minutes after the hour, from 06:30 to 20:30, resulting in 390 images collected on March 29, 2022. The Wingst dataset, spanning several days, reflects changing weather conditions by selecting series of images from different days. This results into 750 images between 06:30 and 21:30.

This process gives a dataset of 1,140 images captured under various environmental conditions, with 952 images containing at least one wolf. These images were manually labeled with bounding boxes for individual wolves and used as our reference dataset. The objective it, that \shadowwolf ideally detects all animals in our dataset as accurately as the manual annotations.

We used the IPTC IIM standard to store additional metadata (e.g., keywords) within the image files. The open-source software digiKam\footnote{\url{https://www.digikam.org/}} was used to assign weather and light-related keywords: \textit{rain}, \textit{snow}, \textit{fog}, \textit{sunny}, \textit{twilight}, \textit{night}, \textit{overcast}, and \textit{contre-jour}. Each image could have multiple keywords. As we will later analyze detection performance concerning lighting conditions in Section~\ref{sec:performance}, we focus on this predefined list of terms.

Additionally, we noted whether images were captured during the day or at night. Table~\ref{tab:image_properties} lists the number of images for each property and their corresponding metadata sources.

\begin{table}
    \centering
    \begin{tabulary}{\columnwidth}{ R C C }
        \textbf{Property} & \textbf{Data Source} & \textbf{Count}\\
    \hline
        Day & Image & 806\\
    \hline
        Night & Image & 334\\
    \hline
        Overcast & IPTC & 389\\
    \hline
        Sunny & IPTC & 357\\
    \hline
        Night & IPTC & 334\\
    \hline
        Twilight & IPTC & 18\\
    \hline
        Rain & IPTC & 50\\
    \hline
        Contre-jour & IPTC & 0\\
    \hline
        Snow & IPTC & 0\\
    \hline
        Fog & IPTC & 0\\
    \hline\hline
        Total & & 1140\\
    \hline
    
\end{tabulary}
    \caption{The number of images with a given property and data source. One image can have multiple IPTC keywords. Therefore, those do not sum up to the total number of images.}
    \label{tab:image_properties}
\end{table}

\subsection{Object Detection Models and Frameworks}
\label{sec:frameworks}
Analysis of images and object detection are two main applications of computer vision. \shadowwolf utilizes this technology to detect objects and create labels. Our goal is not to develop an entirely new architecture for wolf detection but to leverage and adapt existing ones while remaining model-independent. The architecture and model selection for deployment will depend on the user's requirements: whether it is for an embedded system or a GPU server, for real-time detection or preprocessing for human evaluation. We aim to be as application and model-agnostic as possible.

For object detection in \shadowwolf, we evaluated state-of-the-art libraries and frameworks and selected the most suitable one for our evaluation. This section briefly summarizes our findings.

Three classes of algorithms can be distinguished. 
First, \textbf{Classification} algorithms analyze the entire image and return a list of classes along with the probability of each class being present in the image. However, these algorithms do not provide the position of the objects within the image.
Secondly, when classification is combined with object location, we get an \textbf{Object Detection} algorithm. This type of algorithm provides a bounding box and a probability that the object within this box belongs to a particular class.
Thirdly, \textbf{Image Segmentation} takes object detection a step further by adding pixel-accurate boundaries around the objects instead of a simple bounding boxes. This technique provides a detailed outline of each object, capturing its precise shape and contours within the image.

To train a model for object detection, a dataset with labeled objects is required. This is called annotation and can be done in three ways: 1) labeling individual images with a single dominant class, 2) marking objects with bounding boxes in an image, or 3) creating pixel-wise masks for image segmentation. Object detection typically uses one of the first two methods, while image segmentation relies on pixel-wise masks. For \shadowwolf, we focus on object detection, which requires bounding boxes or image segments with corresponding classes.

Training a model from scratch can be resource-intensive in terms of time, computing power, and energy. To address this, transfer learning has become a common practice. This involves using a pre-trained model, originally trained on a different set of classes, and only retraining the final layers to adapt it to new classes or domains. The lower layers, which extract fundamental features, remain unchanged. This approach significantly accelerates development by leveraging existing models and focusing on adapting only the final layers.

Several frameworks and ecosystems for machine learning are available. Here, we provide a brief introduction, with a detailed evaluation in \ref{app:frameworks}.

One of the most well-known frameworks is \textbf{TensorFlow} \cite{tensorflow2015-whitepaper}, developed by Google. It supports a range of devices, including embedded systems and through a JavaScript implementation. TensorFlow's \textit{Model Garden} provides various models adaptable to different machine learning tasks. Comprehensive documentation and tutorials are available on its website\footnote{\url{https://www.tensorflow.org/}}. TensorFlow primarily focuses on developing new models and deploying them across various devices.

\textbf{PyTorch} \cite{NEURIPS2019_9015}, developed by META AI (formerly Facebook), is another prominent machine-learning framework. It is widely favored in research for its clean structure and intuitive design, facilitating model training and evaluation. Like TensorFlow, PyTorch also provides numerous models and pre-trained weights.

\textbf{YOLO} (You Only Look Once) \cite{7780460} is an object detection algorithm with a toolchain based on PyTorch that simplifies training and evaluation. It achieves excellent results with minimal effort and scales easily across various hardware platforms. Unlike TensorFlow and PyTorch, which require manual setup for image pre-processing and data augmentation, YOLO provides built-in support for these steps.

YOLO offers models with varying image resolutions and sizes, denoted by a letter following the version number. For instance, YOLOv5n is the nano model, YOLOv5s is small, and YOLOv5l is large. Model size affects training time, model size, and execution speed. Default models use 640 px images, with higher-resolution options (1280 px) available, marked with a six (e.g., YOLOv5n6, YOLOv5s6, YOLOv5l6). Details on the pre-trained models can be found online\footnote{\url{https://github.com/ultralytics/yolov5#pretrained-checkpoints},\\ accessed 2025-02-20}.

Our evaluation of YOLO on our dataset, as shown in \ref{app:yolo_details}, yielded significantly higher true positives and fewer false positives compared to models from PyTorch and TensorFlow. YOLO’s integrated pre-processing and augmentation streamline model optimization, and its varied model sizes facilitate deployment on different hardware, including embedded systems.

\textbf{MegaDetector} \cite{beery2019efficient} is designed to detect animals, people, and vehicles in camera trap images. Unlike the previously mentioned frameworks, MegaDetector identifies the presence of animals but does not classify them by species. It is based on YOLO, with earlier versions using TensorFlow.
For \shadowwolf, MegaDetector's bounding boxes can be used in the segmentation phase. However, due to its long run time and large model size, we opted for faster classic background subtraction methods.

Frameworks like PyTorch and TensorFlow are optimized to develop new architectures and models for all kinds of machine-learning tasks. They offer many tools and tutorials to generate and evaluate new models but also require a deeper understanding of their structure and the input parameters. MegaDetector detects general animals and no species. YOLO focuses on object detection and helps to get good results quickly. It is scalable to run on different machines and performs well on our dataset. Therefore, we decided to continue this work using YOLO.

\subsection{Metrics and Evaluation}
\label{sec:metrics}

\begin{figure}
    \centering
      \includegraphics[width=0.4\columnwidth]{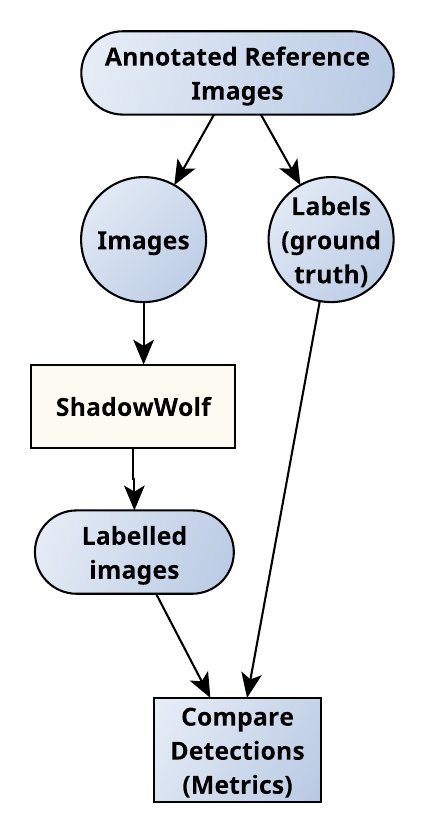}
      \caption{Evaluation process of \shadowwolf: The automatically generated labels are compared against user-provided ground truth. Ideally, the labeled images would match the ground truth with 100\% accuracy, indicating that \shadowwolf performs as effectively as manual labeling.}
      \label{fig:eval_short}
\end{figure}

To evaluate \shadowwolf, we use the reference dataset of 1140 manually labeled images as described in Subsection~\ref{sec:datacollection}. We input these images (without labels) into \shadowwolf to generate predictions. The manually labeled images serve as the ground truth. We then compare the generated labels against the ground truth using our metrics. Ideally, the labels should match perfectly. The evaluation process is illustrated in Figure~\ref{fig:eval_short}.

To compare the bounding boxes from the ground truth with the detections, we use \textbf{Intersection over Union} (IoU), also known as the Jaccard index, as described by Equation~\eqref{eq:iou}. The IoU score ranges between 0 (no overlap) and 1 (complete overlap). A threshold value of \(\alpha=0.5\) is commonly used. If \(J(A,B) \ge \alpha\), the detection is considered true positive.

\begin{equation}
J(A,B) = \frac{|A \cap B|}{|A \cup B|}
    \label{eq:iou}
\end{equation}

The used metrics and their meanings are

\begin{itemize}
    \item \textbf{True Positive (TP):}  A detection that correctly identifies an element as the corresponding class, with \(J(A,B) \ge \alpha\).
    \item \textbf{False Positive (FP):} A detection that incorrectly identifies an element as belonging to a class when it does not, with\ \(J(A,B) < \alpha\).
    \item \textbf{False Negative (FN):} An instance that is not detected by the algorithm.
    \item \textbf{True Negative (TN):} Not applicable in object detection tasks, as not every part of the image belongs to a class.
    \item \textbf{Precision:} Reflects the accuracy of the predictions, representing the percentage of correct detections. It is calculated as shown in Equation~\eqref{eq:precision}.
    \item \textbf{Recall:} Measures whether the algorithm has found all instances of a class. The calculation is depicted in Equation~\eqref{eq:recall}.
    \item \(\textbf{F}_\textbf{1}\)\textbf{-Score:} The harmonic mean of Precision and Recall. A perfect model achieves an \fone-Score of 1, indicating all instances are detected with no false positives. It is calculated as shown in Equation~\eqref{eq:f1}. The goal is to achieve an \fone-Score as close to 1 as possible.
\end{itemize}

\begin{equation}
    \text{precision} = \frac{\text{TP}}{\text{TP}+\text{FP}}
    \label{eq:precision}
\end{equation}

\begin{equation}
    \text{recall} = \frac{\text{TP}}{\text{TP}+\text{FN}}
    \label{eq:recall}
\end{equation}

\begin{equation}
    \text{F}_1 = 2 \cdot \frac{\text{precision} \cdot \text{recall}}{\text{precision} + \text{recall}}
    \label{eq:f1}
\end{equation}

This work primarily uses Precision, Recall, and the \fone-Score as the main metrics. We also report the absolute numbers of True Positives (TP), False Positives (FP), and False Negatives (FN). The (mean) average precision (AP/mAP), which calculates precision across all classes, is not used here since most of our dataset consists of a single class (wolf). We focus on Precision and Recall specifically for this class and compute the \fone-Score.

\section{System Architecture}
\label{sec:architecture}
In Section~\ref{sec:sota}, we evaluated various projects related to our work. Most focus on developing or optimizing models. In contrast, \shadowwolf adopts a different approach by concentrating on the training and evaluation of arbitrary models. The core idea is to integrate state-of-the-art technologies for image preprocessing and data handling into a cohesive workflow, enhanced by crowd-assisted validation. \shadowwolf emphasizes repeatability and reproducibility by packaging results, including configuration and extensive log files, into a single output directory. Users only need to adjust minor parameters, with all major modifications made in the configuration file, thus improving documentation.

Additionally, \shadowwolf is designed to support continuous training environments, where the training dataset can be progressively improved through labeling camera trap images with minimal administrative effort.

\begin{figure*}
  \centering
    \includegraphics[width=1\textwidth]{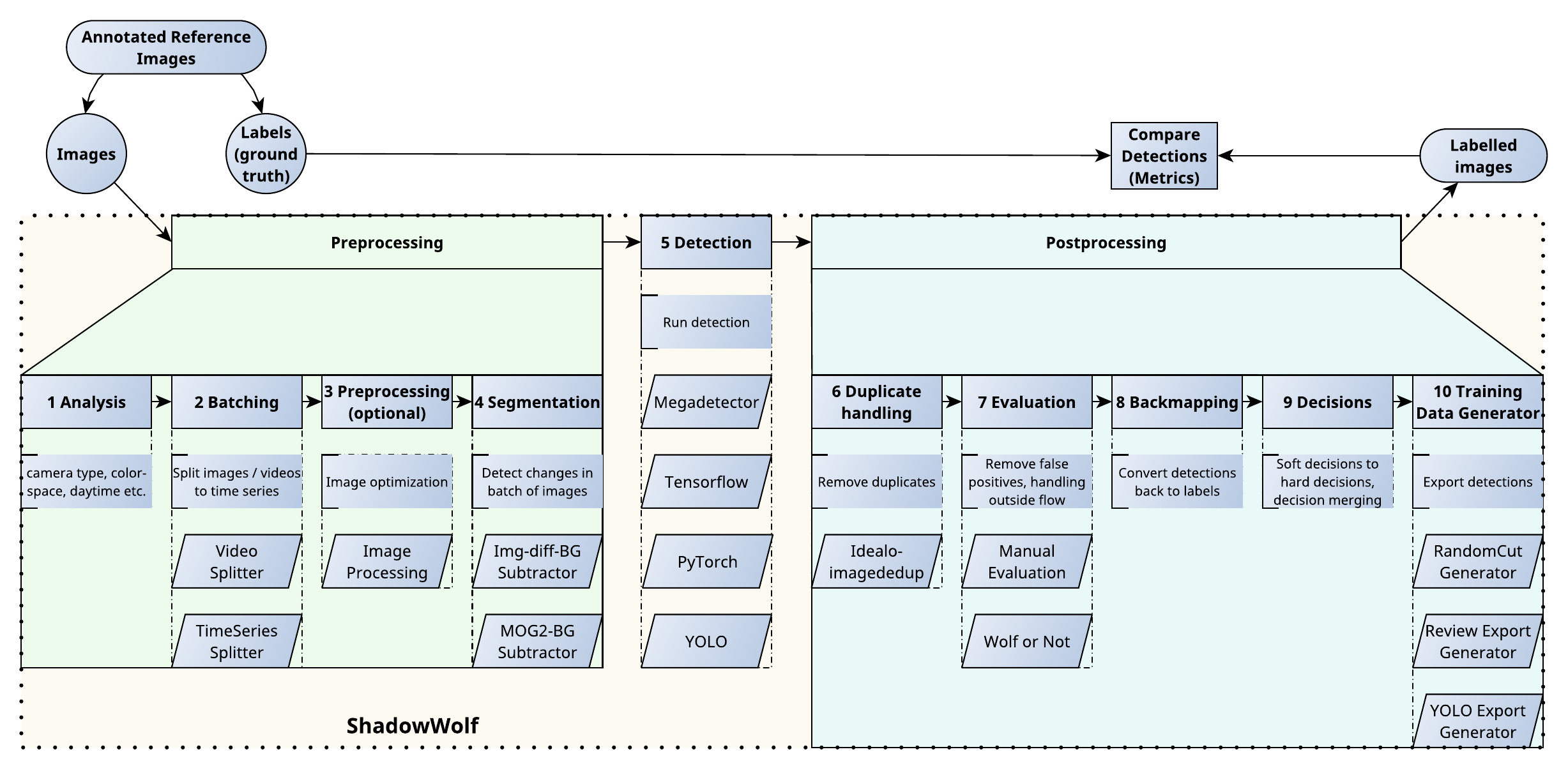}
    \caption{The \shadowwolf framework and its evaluation process aim to transform input videos, images, or image series into well-labeled datasets. The components within the dotted box represent \shadowwolf's three main stages: Preprocessing, Detection, and Postprocessing, each with configurable submodules. The Detection phase utilizes a trained animal detection model that is iteratively improved, as depicted in Figure~\ref{fig:cont_learning}.\newline
    Performance evaluation is conducted using a reference dataset and predefined metrics, illustrated by the boxes outside the dotted area. This involves comparing \shadowwolf-generated labels with manually created ground truth labels. Ideally, these labels should align perfectly. Combined with the iterative model training, as shown in Figure~\ref{fig:cont_learning} this results in automatically generated, enhanced models.\newline
    The framework operates fully automatically and requires minimal administrative intervention.
    }
    \label{fig:flowchart_overview}
\end{figure*}

To achieve this, \shadowwolf operates through several modular steps that can be connected in various configurations. Figure~\ref{fig:flowchart_overview} illustrates the main structure and possible configurations, with \shadowwolf enclosed in the dotted box and the performance evaluation steps from Figure~\ref{fig:eval_short} shown outside.

The framework allows flexible connections between these steps, with each step building on previous information. A central configuration file manages the process, enabling quick re-runs and documentation. Statuses and files are saved after each step, allowing \shadowwolf to restart from any point if errors occur.

The following subsections, we will step through the \shadowwolf framework, while more technical details are provided in \ref{app:shadowwolf_details}.

\subsection{The ShadowWolf Architecture}
We divide \shadowwolf into three main parts: preprocessing, detection, and postprocessing. Preprocessing includes four modules, detection has one, and postprocessing consists of five steps. Each step can implement different alternative methods, e.g., the detection step \circled{5} can use TensorFlow, PyTorch, YOLO or other detection models that are suitable of will become available in the future.

\ref{app:shadowwolf_details} provides detailed descriptions. Here, we offer an overview of the main parts and their key functions. The modular system is adaptable and will evolve based on application needs and future technical (ML) advancements.

\subsubsection{Preprocessing}
The first phase is preprocessing. In step \circled{1}, we analyze the input image files and extract all available metadata and file system information to preserve them for the consequent steps. This information is the image color space, date and time it was taken, camera type etc.

Step \circled{2} focuses on data unification, which involves splitting videos into individual frames or analyzing time correlations within a sequence of images. After this step, we have pure images with, if applicable, marked time-correlation.

Step \circled{3} involves optimizing and adapting these images. For example, spherical distortions from wide-angle lenses can be corrected or the color space can be unified. The goal of this step is to process and standardize the images, ensuring consistency even if they were captured using different cameras or lenses.

The final step in the preprocessing phase is step \circled{4}, segmentation. The object of interest may not always be the dominant part of the image; for instance, animals might appear at the edge of an image. In this step, various techniques are employed to differentiate the object of interest from the background. The segmented objects are then isolated from the background and prepared for detection. This approach ensures that the detection algorithm in the subsequent phase receives input with the highest possible resolution: We focus only on the part of the image not identified as the background. For that, we use MOG2-BG \cite{zivkovic2006efficient,zivkovic2004improved}.

\subsubsection{Detection}
In step \circled{5}, we perform the actual object detection. We designed the interfaces to be as generic as possible, supporting major frameworks such as TensorFlow and PyTorch. In our specific scenario, we use our self-trained YOLO model. The evaluation of the different models is discussed in \ref{app:frameworks}. The selection of YOLO was based on extensive tests and comparative studies, detailed in \ref{app:yolo_details}.

The output of this step typically consists of soft decisions: a list of probabilities indicating the presence of objects within the image or image segment and, depending on the algorithm, a corresponding bounding box. The subsequent postprocessing steps condense this information to produce the final required output.

\subsubsection{Postprocessing}
At this stage, we have images and / or image segments with their respective detections. Our observations on real-world captured images show that similar images with similar detections appear frequently. Those can be caused by barely moving animals or false positive detections on the background. In step \circled{6}, we compare these detections and identify similar objects to reduce the number of images for the subsequent steps: Similar images are handled as one. Later, in step \circled{8}, we undo this step.

\begin{figure}
    \centering
    \includegraphics[width=0.6\columnwidth]{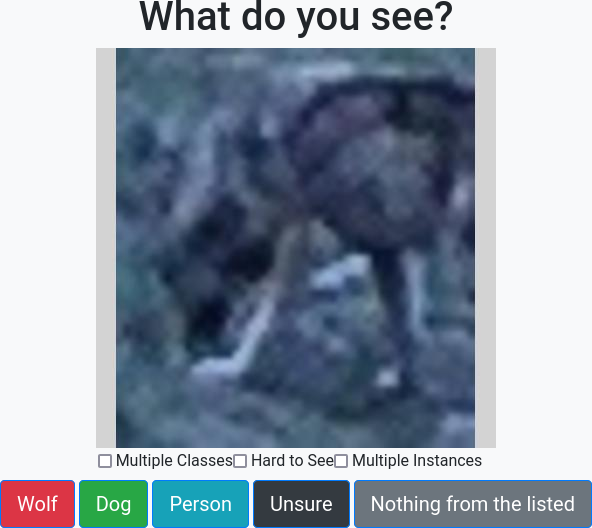}
    \caption{Screenshot from the \wolfornot service: Users view an image and select the corresponding class by clicking on one of the options below the image. Afterwards, the next image is loaded for review.}
    \label{fig:simplelabel_screenshot}
\end{figure}

Step \circled{7} is crucial for enhancing overall performance and reducing the workload for domain experts. Instead of limiting image review to a small group of experts, we incorporate a form of gamification and crowdsourcing by engaging the public. The images are uploaded to our \wolfornot web service, where users can participate by voting on image content. The interface, shown in Figure~\ref{fig:simplelabel_screenshot}, is streamlined and asks the simple question, ``\textit{What do you see?}'' Users select the corresponding option, and the next image is automatically loaded for review. This approach enables volunteers to label large datasets efficiently. To ensure data reliability and prevent misuse, a minimum number of votes is required for each image before continuing with further processing in \shadowwolf. Detailed information about \wolfornot can be found in our previous work in \cite{dede23-animals-wild}.

At this point, we have multiple sources of detections: the probabilities generated by the detection step \circled{5}, the user votes from \circled{7}, and overlapping image areas resulting from nearby detections. In step \circled{8}, we map all detections back to the original image, retaining the bounding boxes and probability scores.

Step \circled{9} involves merging the results from previous steps to form a consolidated decision. For instance, if the detection algorithm assigns a 60\% probability to a region containing a wolf and 100\% of \wolfornot users confirm the presence of a wolf, this reinforces the detection's reliability. We combine these results to produce a final decision, using crowdsourced votes from \wolfornot to support or challenge the initial detection from step \circled{5}. For this step, we also implemented a weighting for the individual results to consider the trust in the different sources.

The final step \circled{10} produces the required output from the processed and now labeled data. This output can take different formats:
\begin{itemize}
  \item We generate a labeled training dataset with full resoultion images that can be used to refine the model in the iterative process descibed in Section~\ref{sec:objectives}.
  \item Another option is to generate a review as required by farmers and associations documenting occurence of animals.
  \item Only the detection with the object of interest are exported for example to train a classifier.
\end{itemize}

\section{Performance Evaluation}
\label{sec:performance}

In this section, we evaluate the performance of our \shadowwolf toolchain using our wolves datasets. However, the toolchain is not limited to wolves in any way and can be adapted to analyze any type of animal in their natural habitats. Our evaluation focuses on the overall detection performance, with a detailed analysis of the impact of daytime on detection accuracy. Additionally, we assess the computational overhead introduced by the toolchain.

\subsection{Use Case}
\label{subsec:use_case}

We applied the \shadowwolf toolchain to our reference dataset and used the configuration listed in Table~\ref{tab:configuration_performance}. For further details, refer to Figure~\ref{fig:flowchart_overview}, Section~\ref{sec:architecture} and \ref{app:shadowwolf_details}.

\begin{table}
    \centering
    \begin{tabularx}{\columnwidth}{ l X }
        \textbf{Step} & \textbf{Settings}\\
    \hline
    \circled{1} Analysis&Extract all available metadata, including IPTC tags and timestamps\\
    \hline
    \circled{2} Batching&We create batches of time-correlated images\\
    \hline
    \circled{3} Preprocessing&No preprocessing was applied as the model was trained on unmodified images\\
    \hline
    \circled{4} Segmentation&Evaluate the effect of segmentation by comparing results with and without the \textit{MOG2-BG-Subtractor}\\
    \hline
    \circled{5} Detection&We use our YOLOv5 model, trained with 1987 images to detect wolves.\\
    \hline
    \circled{6} Duplicate Handling&Implement deduplication to reduce workload in the evaluation phase\\
    \hline
    \circled{7} Evaluation&Use the \wolfornot tool to further assess the detection results\\
    \hline
    \circled{8} Backmapping&Map detections back to the original images\\
    \hline
    \circled{9} Decision&Combine decisions from Detection and Evaluation using a weighted arithmetic mean (0.6 for Evaluation, 0.4 for Detection) to reflect greater trust in human evaluation\\
    \hline
    \circled{10} Training Data Generator&Generate training data in YOLO format. Produce both absolute and soft decision outputs for deeper insight into the workflow. Use a threshold of 0.5 for the absolute decisions\\
    \hline
\end{tabularx}
    \caption{Configuration settings for each step used during the performance evaluation.}
    \label{tab:configuration_performance}
\end{table}

We conducted three main experiments. The first is the \textbf{YOLO Evaluation}, which serves as the baseline. In this experiment, \shadowwolf functionality is disabled to assess the standalone performance of our YOLO model on the reference dataset. Next, in the \textbf{Segmentation Evaluation}, we investigate the impact of segmenting original images into smaller sections, exploring whether this improves the performance of the plain YOLO model.

The final experiment is the \textbf{Complete ShadowWolf Evaluation}, which includes the full workflow with crowdsourcing through our \wolfornot web service. This step demonstrates the overall performance gain achieved by the complete \shadowwolf process still using the same YOLO model from the first two experiments.

Ideally, the performance, measured by the \fone score, should improve with each experiment, demonstrating how \shadowwolf enhances the performance of an existing and unchanged model. The next section presents an evaluation of these experiments and discusses the results based on the metrics outlined in Subsection~\ref{sec:metrics}.

\subsection{General Results}
\label{subsec:results}
First, we evaluated the complete dataset across the three experiments, as detailed in Table~\ref{tab:full_results}. This analysis encompasses all 1140 images, including both day and night shots. The performance of YOLO alone and YOLO with segmentation is nearly identical, with both achieving almost the same \fone-score. Minor variations are observed in false positives, true positives, and false negatives, which aligns with expectations given that YOLO already incorporates segmentation techniques internally.

In the case of \shadowwolf, the generated bounding boxes may not always perfectly align with the objects, leading to an increased number of false positives. Details are given in \ref{app:shadowwolf_details} and Figure~\ref{fig:wolves_labelimg}. To address this, we assessed the results using different IoU thresholds \(\alpha\) changing the value when two overlapping boxes are counted as one. In addition to the default \(\alpha = 0.5\), we tested \(\alpha = 0.25\) and \(\alpha = 0.1\). Table~\ref{tab:full_results} confirms that lowering \(\alpha\) effectively reduces the number of false positives. 

\begin{table}
    \centering
    \begin{tabulary}{\textwidth}{ R C C C C C C }
        &\textbf{FP} & \textbf{TP} & \textbf{FN} & \textbf{Prec.} & \textbf{Recall} & \textbf{F1} \\
    \hline
        \textbf{YOLO} & 81 & 846 & 643 & 0.904 & 0.568 & 0.700\\
    \hline
        \textbf{YOLO \& Segm.} & 90 & 846 & 643 & 0.904 & 0.568 & 0.698\\
    \hline
        \textbf{Full flow}, \(\alpha=0.5\) & 204 & 808 & 681 & 0.798 & 0.543 & 0.646\\
    \hline
        \textbf{Full flow}, \(\alpha=0.25\) & 47 & 965 & 524 & 0.953 & 0.648 & 0.772\\
    \hline
        \textbf{Full flow}, \(\alpha=0.1\) & 26 & 986 & 503 & 0.974 & 0.662 & 0.788\\
    \hline
\end{tabulary}
    \caption{The results from the three runs over the complete dataset containing 1140 images. \(\alpha\) is the decision value for the intersection over union (IoU). If this value is larger than the given \(\alpha\), the detection is considered correct.}
    \label{tab:full_results}
\end{table}

\begin{figure*}
\centering
     \begin{subfigure}[t]{0.24\textwidth}
         \centering
         \includegraphics[width=\textwidth]{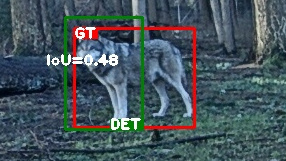}
         \caption{Only half of the wolf was detected during daytime.}
         \label{fig:half_wolf_day}
     \end{subfigure}
     \hfill
     \begin{subfigure}[t]{0.24\textwidth}
         \centering
         \includegraphics[width=\textwidth]{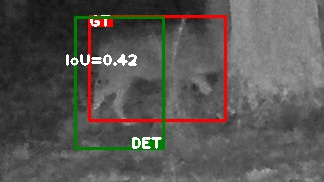}
         \caption{A wolf disappears and is only partly detected.}
         \label{fig:half_wolf_night}
     \end{subfigure}
     \hfill
     \begin{subfigure}[t]{0.24\textwidth}
         \centering
         \includegraphics[width=\textwidth]{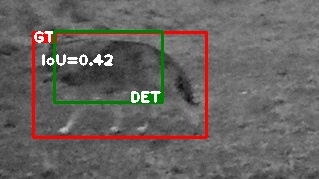}
         \caption{Only a part of the wolf is detected during the night.}
         \label{fig:subwolf_night}
     \end{subfigure}
     \hfill
     \begin{subfigure}[t]{0.24\textwidth}
         \centering
         \includegraphics[width=\textwidth]{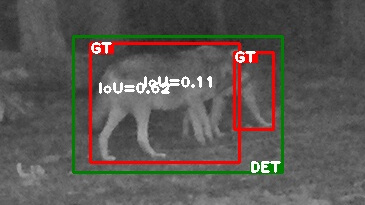}
         \caption{Two wolves were detected as one}
         \label{fig:two_wolves}
     \end{subfigure}
        \caption{Several detections resulting in an IoU smaller than 0.5. The green boxes are the detections, and the red ones are the ground truth. The IoU is given for each set of two boxes.}
        \label{fig:bad_iou}
\end{figure*}

To better understand the impact of the IoU threshold \(\alpha\), we visualized the detections alongside the ground truth and IoU values, as shown in Figure~\ref{fig:bad_iou}. Figures~\ref{fig:half_wolf_day} and \ref{fig:subwolf_night} illustrate cases where only partial detections of wolves occur. This often happens when the wolf is barely moving and goes undetected by YOLO, causing the motion detector to capture only fragments of the animal and resulting in undersized bounding boxes that miss critical details. Figure~\ref{fig:half_wolf_night} highlights a similar issue where a wolf partially obscured by a tree is detected only in its visible portion. Figure~\ref{fig:two_wolves} depicts a scenario where two wolves are closely positioned, with \shadowwolf detecting only one. The ground truth identifies the second wolf separately. In these situations, the boxes are primarily generated by the background subtractor rather than the detection model. Although these boxes are generally less precise than those from the machine learning model, they provide valuable information when detection is otherwise incomplete. Consequently, lowering \(\alpha\)  is justified to accommodate these imperfections.

\subsection{Effect of the Daytime}
\label{subsec:night_images}

Good lighting conditions are crucial for obtaining high-quality detections. To assess the impact of ambient light, we used the data from the analysis module \circled{1} to filter the detections accordingly. Table~\ref{tab:results_daytime} displays the metrics for the 806 daytime images, while Table~\ref{tab:results_nighttime} shows the metrics for the 334 nighttime images.

\begin{table}
    \centering
    \begin{tabulary}{\textwidth}{ R C C C C C C }
        &\textbf{FP} & \textbf{TP} & \textbf{FN} & \textbf{Prec.} & \textbf{Recall} & \textbf{F1} \\
    \hline
        \textbf{YOLO} & 72 & 560 & 538 & 0.886 & 0.510 & 0.647\\
    \hline
        \textbf{YOLO \& Segm.} & 76 & 562 & 536 & 0.881 & 0.512 & 0.647\\
    \hline
        \textbf{Full flow}, \(\alpha=0.5\) & 138 & 571 & 527 & 0.805 & 0.520 & 0.632\\
    \hline
        \textbf{Full flow}, \(\alpha=0.25\) & 23 & 686 & 412 & 0.968 & 0.625& 0.760\\
    \hline
        \textbf{Full flow}, \(\alpha=0.1\) & 8 & 701 & 397 & 0.989 & 0.638 & 0.776\\
    \hline
\end{tabulary}
    \caption{The results from the three runs over the 806 daytime images. The reduced number of total images has to be considered for the false positives, true positives and false negatives.}
    \label{tab:results_daytime}
\end{table}

Comparing the results from daytime images with the overall results reveals a general decline in all metrics, which is unexpected given that good lighting conditions were assumed to enhance detection quality. This decline in precision, recall, and \fone-score during the daytime can be attributed to the fact that while humans can easily identify distant wolves, particularly when annotating time series data, the detector struggles with such tasks, leading to poorer results.

Conversely, our YOLO model appears to have difficulty with wolves that are only partially visible, such as those entering or leaving the camera's field of view. In these cases, the human interaction provided by \wolfornot is crucial for identifying these wolves, thereby improving overall detection performance.

For the 334 nighttime images, as shown in Table~\ref{tab:results_nighttime}, the \fone-score generally improves across all experiments. This improvement is due to the limited detection range of the camera’s infrared spot, which can illuminate only up to 15 meters according to the datasheet. Consequently, the advantage of human annotation for distant wolves disappears, leading to higher \fone-scores. Additionally, the benefits of motion detection for distant wolves in \shadowwolf are reduced during nighttime, making the performance of pure YOLO approaches more comparable. For this dataset, reducing the \(\alpha\) value for IoU further enhances the metrics.

\begin{table}
    \centering
    \begin{tabulary}{\textwidth}{ R C C C C C C }
        &\textbf{FP} & \textbf{TP} & \textbf{FN} & \textbf{Prec.} & \textbf{Recall} & \textbf{F1} \\
    \hline
        \textbf{YOLO} & 9 & 286 & 105 & 0.969 & 0.731 & 0.834\\
    \hline
        \textbf{YOLO \& Segm.} & 14 & 284 & 107 & 0.953 & 0.725 & 0.824\\
    \hline
        \textbf{Full flow}, \(\alpha=0.5\) & 66 & 237 & 154 & 0.782 & 0.606 & 0.683\\
    \hline
        \textbf{Full flow}, \(\alpha=0.25\) & 24 & 279 & 112 & 0.921 & 0.714 & 0.804\\
    \hline
        \textbf{Full flow}, \(\alpha=0.1\) & 18 & 285 & 106 & 0.941 & 0.729 & 0.821\\
    \hline
\end{tabulary}
    \caption{The results from the three runs over the 334 nighttime images. Again, the reduced number of total images has to be considered.}
    \label{tab:results_nighttime}
\end{table}

\subsection{Required Computing Capabilities}
\label{subsec:computing}
We also assessed the computing resources used in our experiments, which were conducted on a standard computer with an Intel i7 processor (four cores) and 12 GB of RAM. The runtimes for each experiment are listed in Table~\ref{tab:runtime}. The \shadowwolf process is divided into two parts: Part 1 involves preparing data for export to the \wolfornot web service, while Part 2 handles processing the data from the web service. The duration of voting on \wolfornot varies with the number of users and their engagement, taking between 1 to 4 days to collect sufficient votes for 2,274 subimages.

The runtimes for "YOLO \& segmentation" and "Part 1 of \shadowwolf" are expected to be similar. However, \shadowwolf benefits from deduplication, which reduces the number of images that need preprocessing by YOLO, cutting runtime by approximately 20 minutes. Processing the images with the YOLO model is the most time-consuming task and could be expedited with a GPU.

In terms of disk space, creating numerous intermediate images for debugging and analysis requires significant storage. The reference dataset, consisting of 1,140 images, occupies 1.7 GB. The data exported to \wolfornot is only 71 MB, while the labeled images also amount to 1.7 GB. The debugging images generated during the process add up to about 7.3 GB, though this can be easily reduced by disabling intermediate image creation.

\begin{table}
    \centering
    \begin{tabulary}{\textwidth}{ R L }
        \textbf{Experiment} & \textbf{Runtime} \\
    \hline
        YOLO only & 30.48 minutes\\
    \hline
        YOLO \& Segm. & 88.03 minutes\\
    \hline
        Full \shadowwolf part 1 & 57.35 minutes\\
        Full \shadowwolf part 2 & 9.22 minutes\\
        Full \shadowwolf complete & 66.58 minutes\\
    \hline
\end{tabulary}
    \caption{Comparison of the processing times on our computer. \shadowwolf is divided into two steps. In between, the evaluation using the web service is done. Here, we can not give a time as it depends on the number of users.}
    \label{tab:runtime}
\end{table}

\section{Discussion}
\label{sec:discussion}

The primary goal of \shadowwolf is to enhance object detection in camera trap images with a focus on dynamic model adaptation and reduced effort in generating new models, training datasets and reviewing images.

Our performance evaluation shows that \shadowwolf improves the detection compared to using a standalone model: The \fone-score improved from 0.70 to 0.79 on our reference dataset when using \shadowwolf versus a baseline YOLO object detection model. The individual components of our framework contribute not only to enhanced detection but also facilitate subsequent analysis and seamless integration into automated training and evaluation workflows. For example, the segmentation step of \shadowwolf identifies objects not belonging to the background, focuses on those objects and enables a more targeted and efficient detection process.

A significant feature of \shadowwolf is its integration with our crowd-based \wolfornot service. Everybody can vote on the detections and, thus, strengthen or weaken the results from the object detection. This step is particularly effective in minimizing misclassifications, thereby refining model accuracy through community-driven feedback.

Distant, barely moving or only partly visible animals are hard to detect. However, \shadowwolf’s flexible, modular architecture allows the incorporation of more sophisticated algorithms capable of detecting slow-moving or partially obscured animals. This adaptability ensures that new advancements in detection technologies can be seamlessly integrated into the framework while based on the available resources.

The intended application of \shadowwolf plays a role in determining its operational approach. For generating a new, labeled dataset, higher accuracy is essential, whereas visual inspection support can tolerate greater variability. For extensive datasets, \shadowwolf can substantially decrease the workload by filtering detections to display only those within a specified probability range. This capability also includes automatic handling of detections, discarding results with very high or low probabilities while flagging ambiguous cases (e.g., "Maybe a wolf?") for manual review by domain experts. This targeted approach becomes especially advantageous as the frequency of wolf sightings increases, requiring efficient allocation of expert resources.

\shadowwolf's modularity and extensibility are designed to minimize user interaction. The system can be initialized with a dataset, such as camera trap images, and run autonomously. Integration with \wolfornot is straightforward, leveraging existing APIs for image uploads and result retrieval, enabling users to obtain an annotated dataset with minimal manual input.

We support iterative improvements: the generated training datasets can be used to train updated model versions, which can be evaluated and benchmarked using \shadowwolf's built-in toolset. Improved models can then be integrated into the framework, enhancing overall performance and supporting a cycle of continuous model refinement. It is also possible to generate models with different complexity and computational requirement that can be used on various camera and detection systems. Depending on the costs and the requirements, the generated models can run on a fully fledged GPU server or an embedded, battery driven camera system in the field.

In terms of resource requirements, \shadowwolf is optimized for operation on standard consumer hardware, with GPU acceleration providing significant reductions in processing time, particularly for the detector model's execution phase. This efficiency is advantageous for cost-sensitive users, such as those engaged in wildlife monitoring, ensuring accessibility and practicality in field applications.

\section{Conclusion}
\label{sec:conclusion}

In this work, we introduced \shadowwolf, a comprehensive toolchain designed to automate the labeling of camera trap images and facilitate the training of new models. \shadowwolf employs a combined approach that enhances state-of-the-art object detection algorithms with user evaluations. We assessed various object detection frameworks and integrated a background subtractor to generate potential detections. These detections, presented to users via our interactive web service, \wolfornot, are combined with machine-generated results to improve accuracy by mitigating false positives and capturing false negatives missed by the initial model. The final output is a labeled dataset suitable for training new models or conducting manual analyses of wildlife occurrences.

Our performance evaluation demonstrated significant improvements with \shadowwolf: the \fone-score for detecting wolves increased from 0.7 to 0.8 across all images and from 0.6 to 0.8 for daytime images when using our framework compared to a standalone object detector. We also examined the impact of lighting conditions and the computing requirements of our approach and the object detection model.

Despite these advancements, our system has limitations: the automatically created bounding boxes do not always achieve perfect alignment with the objects of interest. We discussed additional challenges related to detecting objects in camera trap images.

Overall, \shadowwolf offers a flexible and robust framework for those working with camera trap images:

\begin{itemize}
    \item It enables the detection of partially hidden objects through user interaction and evaluation.
    \item It simplifies the labeling process while simultaneously improving detection accuracy.
    \item It is adaptable, allowing for module replacement and performance comparison.
    \item It supports a continuous workflow, automating the analysis, labeling, training, and evaluation processes.
\end{itemize}

We hope this tool will be valuable to those engaged in wildlife monitoring and camera trap research. We invite interested users to contribute to the project. The source code is available under the GNU General Public License v3.0 on GitHub\footnote{\url{https://github.com/ComNets-Bremen/ShadowWolf}}. Further details and information can also be found in \cite{Dede_2024}.

\subsection*{Further Ideas and Challenges}
\label{subsec:future}
We are actively expanding and refining the components of \shadowwolf. Currently, we are focusing on enhancing the time series evaluation. We plan to introduce a new module that leverages \circled{2}~Batching (c.f.\ Figure~\ref{fig:flowchart_overview}). This module will analyze sequences of images, particularly examining the locations of detections and their associated probabilities over time. By evaluating how detections change or persist, we aim to improve the reliability of these detections, providing better confidence based on temporal consistency.

Another key area of development is the deployment of \shadowwolf. We are exploring methods to automate the transfer of images from camera traps to the system. Challenges include limited cellular Internet availability and data transfer constraints. We are working to develop workflows that accommodate these constraints and meet the needs of potential users.

\section{Acknowledgements}
The authors thank the following parks for supporting us: Alternativer Bärenpark Worbis, Wildpark Lüneburger Heide, Wingster Waldzoo.
This work is part of the \mainzaun project supported by the Federal Ministry of Food and Agriculture, Germany.

\appendix
\section{\break Detailed Steps of ShadowWolf}
\label{app:shadowwolf_details}

The concept and architecture of \shadowwolf were presented in Section~\ref{sec:architecture}, with Figure~\ref{fig:flowchart_overview} providing a comprehensive overview. In this appendix, we will detail the ten individual steps and the currently implemented functionality of the overall architecture, referring to Figure~\ref{fig:flowchart_overview}. Steps \circled{1} to \circled{4} represent the preprocessing phase, detection is executed in step \circled{5}, and steps \circled{6} to \circled{10} perform postprocessing, as outlined in Section~\ref{sec:architecture}.

\subsection{General Setup and Configuration}
Each step of \shadowwolf is implemented in Python 3.11 (current at the time of this writing). For the complete documentation, including the used library version and access to the source code, visit our GitHub repository\footnote{\url{https://github.com/ComNets-Bremen/ShadowWolf}}.

The configuration is managed via one JSON file with two main sections:

\begin{itemize}
    \item \textbf{General Configuration:} Defines the input directory and file types.
    \item \textbf{Module Configuration:} Lists the modules in order, each corresponding to a step in \shadowwolf, with their specific parameters.
\end{itemize}

An SQLite database is used to track intermediate results and module input and output. Each module has a dedicated directory for storing binary data, such as preprocessed images. For additional details, refer to the documentation and source code on our GitHub repository.

The following subsections describe each individual step, each implemented as a separate Python module.

\subsection{Analysis}

In step \circled{1}, we extract meta information from the original images, including EXIF (Exchangeable Image File Format), IPTC (IPTC Information Interchange Model) data, and image size. This information, such as camera details, timestamps, copyright, keywords, and image descriptions, can get lost during the images processing.

Key attributes we store are:
\begin{itemize}
    \item \textbf{Colorspace:} Important for distinguishing between day and night images and for color-dependent algorithms.
    \item \textbf{Camera Type, Lens Type, and Serial Number:} Useful for identifying and correcting camera-specific distortions.
    \item \textbf{EXIF DateTime Fields:} Used to group images by time, creating series of images; more reliable than file system attributes.
    \item \textbf{IPTC Keywords:} For filtering images during analysis.
    \item \textbf{Original Image Resolution}.
\end{itemize}

\subsection{Batching}
Camera traps often capture a series of images triggered by motion, typically taken at fixed intervals, such as two images per second as in our case. In step \circled{2}, we manage this correlation, which is useful for further processing, as images within the same batch share similar light, weather, and environmental conditions.

We implemented two types of splitters that split the series of images into batches:

\begin{itemize}
    \item \textbf{Time Series Splitter:} Analyzes EXIF timestamps of a series of images usually taken by a camera triggered by a motion detector and splits it into batches. In our case, typically, 10 to 70 images are captured when an object moves through the frame. This splitter organizes these images into separate batches.
    \item \textbf{Video Splitter:} Converts short video sequences into batches of images, similar to the output of the Time Series Splitter.
\end{itemize}

Since most of our data consists of individual images, we focus on the Time Series Splitter. For batching, we use the EXIF tag \textit{DateTime}, which in our case is equivalent to \textit{DateTimeOriginal} and \textit{DateTimeDigitized}. Images are sorted by date and time, and if the time between consecutive images exceeds a defined period (set to 5 seconds in our experiments), \shadowwolf creates a new batch.

\subsection{Preprocessing}
In step \circled{3}, images are optimized to ensure uniformity across different sources, allowing the following steps to focus on object content rather than image variations. Key objectives include:

\begin{itemize}
    \item \textbf{Removing Overlays:} Camera traps may add logos or timestamps, which can distract models from important content, especially when images come from various sources.
    \item \textbf{Correcting Distortions:} Wide-angle lenses can cause spherical distortions, bending straight lines and altering object appearance based on their location in the image.
\end{itemize}

We use OpenCV to correct these distortions based on camera serial numbers, leveraging parameters from the metadata collected in the \circled{1}~Analysis.
Additional optimizations, such as normalizing colorspaces or adjusting brightness, can also be added.

\subsection{Segmentation}
In step \circled{4}, we concentrate on identifying and isolating objects of interest, particularly in scenarios where the target object, such as a wolf, may be obscured by elements like bushes or trees.

Key goals include:
\begin{itemize}
    \item \textbf{Handling Information Loss:} Downscaling high-resolution images can lead to the loss of important details, especially when using lower-resolution models. This step identifies changing regions and extract sub-images for focused analysis in subsequent steps.
    \item \textbf{Enhancing Detection:} By segmenting the image, we can detect objects that might be missed by the model, potentially identifying false negatives. Our app, \wolfornot as used later in step \circled{7}~Evaluation, allows users to review these sub-images to spot hard-to-see objects, improving overall performance.
\end{itemize}

We offer two segmentation options:
\begin{itemize}
    \item \textbf{MOG2-BG Subtractor:} This well-known algorithm uses Gaussian Mixture Models\cite{zivkovic2006efficient,zivkovic2004improved} to separate moving objects from the background, generating bounding boxes around detected movements. It benefits from the \circled{2}~Batching.
    \item \textbf{Img-diff-BG Subtractor:} This method creates an average image from the batch and compares each individual image to this average, marking differences as boxes. This simple approach can effectively detect movements and is based on a technique from pyimagesearch.com\footnote{\url{https://pyimagesearch.com/2015/05/25/basic-motion-detection-and-tracking-with-python-and-opencv/}, accessed 2025-02-20}.
\end{itemize}

Another option for this step is MegaDetector, but we did not implement it due to its high computational resource requirements.

\subsection{Detection}

This step \circled{5} is the core of the framework. As detailed in Section~\ref{sec:frameworks} and \ref{app:frameworks}, we evaluated several frameworks for object detection. Due to its scalability, flexibility, and accuracy, we chose YOLOv5 for this step.

This block outputs soft decisions, providing bounding boxes with detected classes and their probabilities. We focus on the class with the highest probability for subsequent steps, suppressing the others.

\begin{figure*}
\centering
     \begin{subfigure}[c]{0.18\textwidth}
         \centering
         \includegraphics[width=\textwidth]{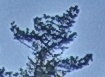}
     \end{subfigure}
     \hfill
     \begin{subfigure}[c]{0.18\textwidth}
         \centering
         \includegraphics[width=\textwidth]{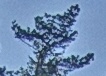}
     \end{subfigure}
     \hfill
     \begin{subfigure}[c]{0.18\textwidth}
         \centering
         \includegraphics[width=\textwidth]{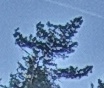}
     \end{subfigure}
     \hfill
     \begin{subfigure}[c]{0.18\textwidth}
         \centering
         \includegraphics[width=\textwidth]{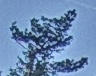}
     \end{subfigure}
     \hfill
     \begin{subfigure}[]{0.18\textwidth}
         \centering
         \includegraphics[width=\textwidth]{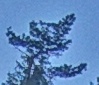}
     \end{subfigure}
        \caption{Five images detected as similar by the image deduplication process are consolidated into a single image, reducing the workload for the following steps. The detections are then applied to all similar images to prevent information loss.}
        \label{fig:similar_images}
\end{figure*}

\subsection{Duplicate Handling}

During our evaluation, we often encountered numerous similar images with identical or near-identical objects. This issue is addressed in the step \circled{6}. For example, similar to Figure~\ref{fig:similar_images}, we detected 62 similar-looking wolves. To reduce the workload and focus on valuable training data, we grouped these images and treated them as a single entity. This reduces the amount of images to be processed in the next step \circled{7}. In the \circled{8}~Backmapping step, detections are extended to all similar images to avoid information loss.

We use the imagededup-Handler, based on \cite{idealods2019imagededup} by idealo, to efficiently detect and group similar images, treating them as one for the next steps.

\subsection{Evaluation}

The step \circled{7} is crucial for enhancing detection quality through user interaction. We offer two options for reviewing detections:

\begin{itemize}
    \item \textbf{Wolf-or-Not:} This web service, implemented by us\footnote{\url{https://github.com/ComNets-Bremen/Wolf-or-Not/}}\cite{dede23-animals-wild}, allows users to review and validate detections, helping to refine the output. Users can flag false positives or confirm detections in a web interface, as shown in Figure~\ref{fig:simplelabel_screenshot}.
    \item \textbf{Manual Evaluation:} This labor-intensive method involves manually validating detections and removing incorrect ones. It is primarily used for debugging.
\end{itemize}

\begin{figure}
    \centering
    \includegraphics[width=0.8\columnwidth]{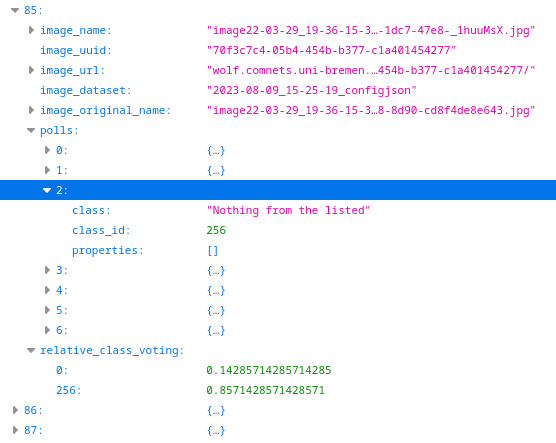}
    \caption{\wolfornot exports evaluation results as a JSON file. The file includes voting data, such as the results from six polls in this example. Here, 86\% of users selected "\textit{Nothing from the listed}" while 14\% identified the image as a "\textit{Wolf}".}
    \label{fig:simplelabel_export}
\end{figure}

For our work, we use \wolfornot due to its efficiency. The service evaluates images from both the Segmentation~\circled{4} and Detection~\circled{5} steps. The images are uploaded to \wolfornot \footnote{Available at \url{https://wolf.comnets.uni-bremen.de}} and displayed to users, who vote on the detected classes. Each image is reviewed by at least three users to ensure reliable results. The outcome is exported as a JSON file, detailing user votes, as shown in Figure~\ref{fig:simplelabel_export}. This helps to identify false positives and improving detection accuracy without the need of well-trained experts.

\subsection{Backmapping}

\begin{figure}
  \centering
    \includegraphics[width=1\columnwidth]{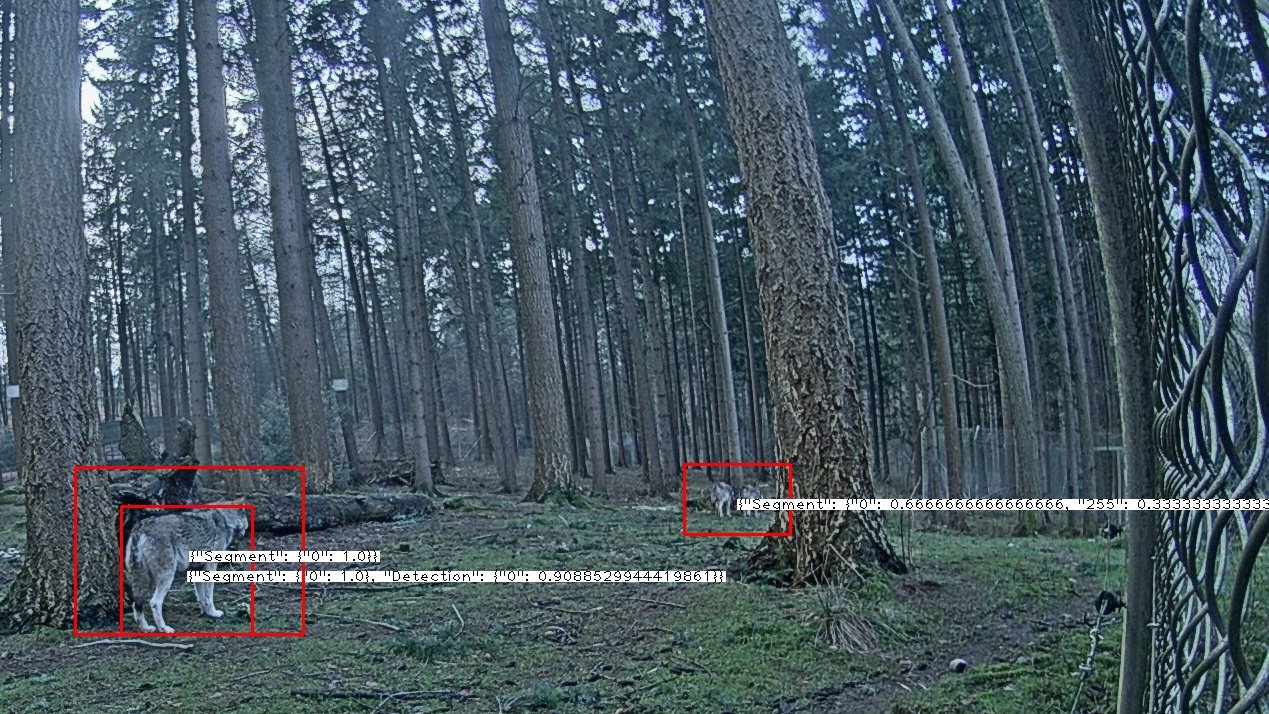}
    \caption{Example output from the backmapping module. For the left wolf, there are two bounding boxes: one from segmentation and one from detection. Each box includes probabilities from both \wolfornot and the detector. In contrast, the detector did not identify the right wolf. For this instance, only the user-generated probabilities from \wolfornot are available.}
    \label{fig:backmapping_example}
\end{figure}

After segmentation and detection steps, step \circled{8} integrates findings into the original high-resolution images. This step frames detected objects back to their source images, including details such as detection origin, class probability, and bounding boxes. For example, as shown in Figure~\ref{fig:backmapping_example}, the left wolf has two bounding boxes: an outer one from segmentation and an inner one from YOLO detection. YOLO was 90.8\% confident in its detection, and users of \wolfornot confirmed the presence of a wolf with 100\% agreement. Conversely, YOLO missed the wolves on the right, although users agreed with 67\% certainty that one was present.

\subsection{Decisions}
In step \circled{9}, we consolidate the information from earlier steps, particularly the \circled{5}~Detection and \circled{7}~Evaluation phases, to calculate a final decision. The process begins with merging overlapping bounding boxes from the Backmapping step, utilizing the Jaccard index (IoU) to determine overlaps (see Figure~\ref{fig:backmapping_example}).

Each bounding box and its detections are combined using a weighted average of probabilities. This method ensures decisions are based on all available data, enhancing accuracy and reliability by adjusting the influence of each detection source according to its confidence level.

\begin{table}
    \centering
\begin{tabular}{ c l }
    \textbf{Symbol} & \textbf{Meaning} \\
    \hline
    \(c\) & The specific class \\
    \(i\) & The detector \\
    \(d_{c,i}\) & The detection from the given detector for a specific class\\
    \(w_i\) & The weight of the given detector\\
    \(\bar{d}_c\) & The result of the object being in a specific class\\
    \hline
\end{tabular}
    \caption{The notations for the weighted probabilities.}
    \label{tab:notations}
\end{table}

For that, we assume that all detectors return the detections in the closed interval \(0-1\) \eqref{eq:interval}.
\begin{equation}
    d_{c, i} \in\mathbb{R} \mid 0 \le d_{c, i} \le 1, \forall c, \forall i
\label{eq:interval}
\end{equation}

The weights for all detectors have to sum up to one to form a valid probability function \eqref{eq:prob_function}

\begin{equation}
    \sum_{\forall i} w_i = 1
\label{eq:prob_function}
\end{equation}

Under the assumptions \eqref{eq:interval} and \eqref{eq:prob_function}, the weighted arithmetic mean can be described as in \eqref{eq:wam}.

\begin{equation}
    \bar{d}_c = \sum_{\forall i} w_i \cdot d_{c, i}
    \label{eq:wam}
\end{equation}

This computes a combined probability for each class by aggregating the results from all detectors. When weights are equal, this is simply the arithmetic mean. If multiple detections originate from the same source, they are used twice and normalized to ensure fairness.

The class with the highest combined probability is selected as the result. This class and its probability are then stored for further processing. The training data generator will convert these probabilities into a format suitable for subsequent use.

\subsection{Training Data Generator}
\begin{figure}
    \centering
      \includegraphics[width=1\columnwidth]{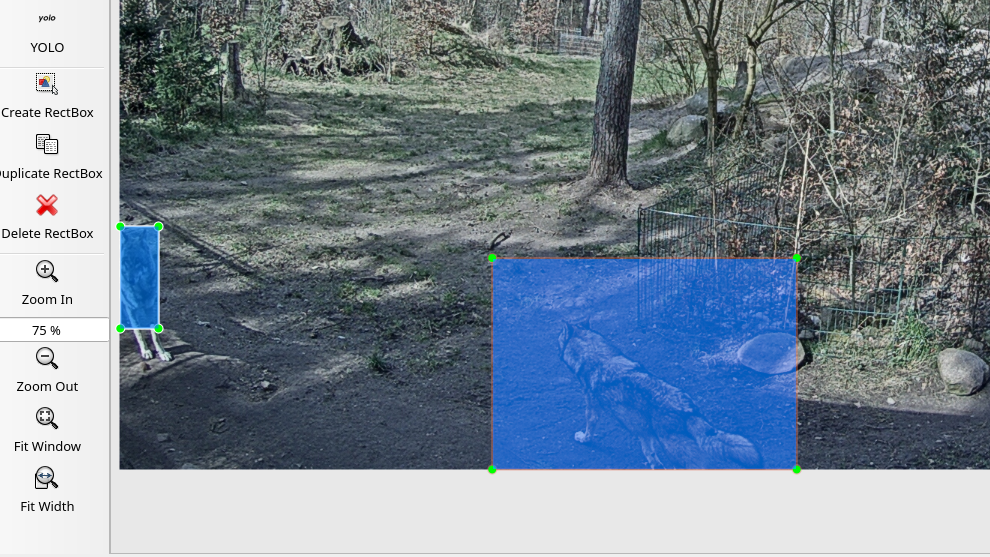}
      \caption{The automatically generated training dataset can be reviewed and refined using tools such as labelImg. This step allows for fine-tuning of bounding boxes to enhance the model's quality. For instance, the left box may need slight extensions, while the right one is too large.}
      \label{fig:wolves_labelimg}
  \end{figure}

In the final step \circled{10}, we finalize the process by providing several options to export and validate the results.
The bounding boxes from previous steps might need adjustments to ensure high-quality training data. This is caused by the use of background subtraction and combining several detections. Tools like labelImg\footnote{\url{https://github.com/HumanSignal/labelImg}, accessed 2025-02-20} can be used to refine these boxes, as shown in Figure~\ref{fig:wolves_labelimg}.

The step offers three export options:

\begin{itemize}
    \item \textbf{YOLO Format:} Export full images and labels with hard decisions for direct training of a YOLO model.
    \item \textbf{Soft Decisions:} Export full images and labels with soft decisions for detailed evaluation.
    \item \textbf{No Output:} Use the previous step’s decisions directly for debugging purposes.
\end{itemize}

Additional output formats can be implemented as needed. For example, detections can be cropped and formatted for training other models, like MobileNet, in TensorFlow or PyTorch.

\section{\break Evaluation of Detection Models}
\label{app:frameworks}
A variety of object detection frameworks are available, and we evaluated several leading ones based on our requirements.

For \textbf{TensorFlow}, we tested the MobileNetV2 architecture \cite{sandler2018mobilenetv2}, an object classification algorithm, on our wolf image dataset. This classifier provides predictions with associated probabilities. MobileNetV2, originally trained on the ImageNet dataset \cite{ILSVRC15}, includes classes like \textit{timber wolf} (ID 269), which theoretically should enable wolf detection.

Initially, we evaluated this model on a Flickr dataset with 289 high-quality images tagged \textit{wolf}. The top detections included \textit{timber wolf} (203), \textit{coyote} (54), and \textit{red wolf} (17). When applied to our dataset of 315 images, the top detections shifted to \textit{coyote} (31), \textit{African hunting dog} (25), and \textit{tabby} (24), with only three images correctly classified as \textit{wolf}. To improve accuracy, we attempted transfer learning by fine-tuning MobileNetV2 on our wolf dataset of 1883 images, but this approach yielded poor results due to the dataset’s imbalance \cite{7727770}. Solutions such as data augmentation or dataset balancing could mitigate these issues, highlighting the importance of pre- and post-processing in our \shadowwolf toolchain.

For \textbf{PyTorch}, we performed a similar evaluation with MobileNetV2 pre-trained on ImageNet. On the Flickr dataset, the model identified \textit{timber wolf} (208), \textit{coyote} (62), and \textit{kit fox} (6). When tested on our wolf dataset, the detections included \textit{grey whale} (64), \textit{velvet} (24), and \textit{snow leopard} (19), with only ten wolves detected correctly. Transfer learning in PyTorch, however, showed promising results for our dataset, correctly identifying all wolves, although it overfitted on the Flickr dataset with only 19 correct detections. This overfitting is typical with small or similar datasets and static camera positions. Improved training datasets, as offered by \shadowwolf, could alleviate this issue. The differences in transfer learning outcomes between TensorFlow and PyTorch are attributed to variations in their approaches to fixed and retrained layers.

\textbf{YOLO} provides various models with different image resolutions and sizes, denoted by suffixes (e.g., YOLOv5n for nano, YOLOv5s for small, YOLOv5l for large). Higher resolutions like 1280~px per side are available in models marked with a six (e.g., YOLOv5n6). YOLO models generally performed better on our dataset, showing higher true positives and fewer false positives compared to PyTorch and TensorFlow models. YOLO’s built-in preprocessing, including image scaling and augmentation, simplifies model optimization. Additionally, YOLO’s model sizes are adaptable for various hardware, making it suitable for deployment on embedded systems. Consequently, YOLO was selected for further evaluation and implementation, as discussed in \ref{app:yolo_details}.

\begin{figure}
    \centering
      \includegraphics[width=\columnwidth]{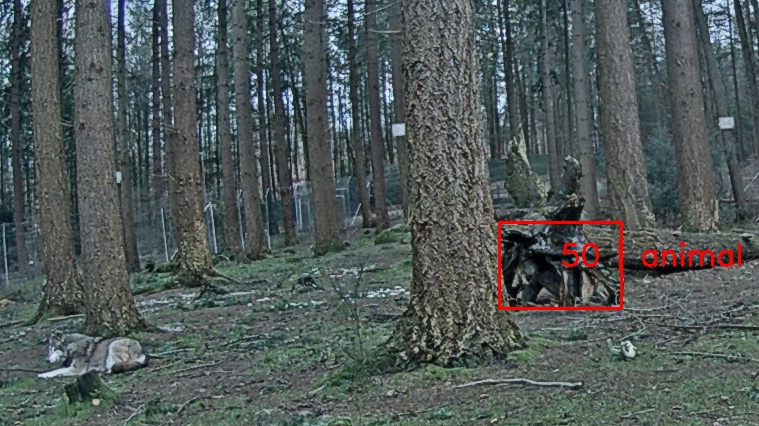}
      \caption{In this example, MegaDetector identified the tree as an animal with 50\% certainty, while the wolf in the lower left corner was not detected.}
      \label{fig:megadetector}
\end{figure}

\textbf{MegaDetector} was also tested on our dataset. Although it generally performed well, it misclassified the tree trunk as a wolf with 50\% certainty and failed to detect the wolf in the lower left corner (see Figure~\ref{fig:megadetector}). MegaDetector's detections are robust but struggle with sitting, lying, or distant wolves. The large model size (267M) also results in slower processing times, requiring 3.3 seconds per image on our server.

Due to its excellent adaptability and performance, we focus this work on the YOLO framework and use our self-trained models.

\section{\break YOLO Implementation Details}
\label{app:yolo_details}

\begin{table*}
    \centering
\begin{tabular}{ r r r r r}
    \textbf{Model} & \textbf{Epochs} & \textbf{Layers} & \textbf{Parameters} & \textbf{Size}\\
    \hline
    \textit{YOLOv5l6} & \textit{285} & \textit{346} & \textit{76,134,048} & \textit{147M}\\
    \hline
    YOLOv5s6 & 400 & 206 & 12,315,904 & 25M\\
    \hline
    YOLOv5n6 & 265 & 206 & 3,091,120 & 6.6M\\
    \hline
\end{tabular}
    \caption{Model type, number of layers and parameters and the number of epochs used for training of three YOLO models with different complexity. The user can select the model depending on the available hardware and needs.\newline
    The model highlighted in \textit{italic} on the first line is the one used in this work.}
    \label{tab:training_yolo_complexity}
\end{table*}

\begin{table*}
    \centering
\begin{tabular}{ r r r}
    &  \multicolumn{2}{c}{\textbf{Time Training}}\\
    \textbf{Model} & \textbf{Total} & \textbf{per Epoch}\\
    \hline
    \textit{YOLOv5l6} & \textit{8.2 hrs} & \textit{1.7 mins} \\
    \hline
    YOLOv5s6 & 10.1 hrs & 1.5 mins\\
    \hline
    YOLOv5n6 & 6.6 hrs & 1.5 mins\\
    \hline
\end{tabular}
    \caption{Training time of the different YOLO models on our GPU server.\newline
    The model highlighted in \textit{italic} on the first line is the one used in this work.}
    \label{tab:training_yolo_training}
\end{table*}

\begin{table*}
    \centering
\begin{tabular}{ r r r r r}
    & \multicolumn{4}{c}{\textbf{Time Inference per Image}}\\
    \textbf{Model} &\textbf{RaspberryPi} & \textbf{PC} & \textbf{Server}&\textbf{GPU}\\
    \hline
    \textit{YOLOv5l6} & \textit{4961 ms} & \textit{863 ms} & \textit{163 ms}& \textit{42 ms}\\
    \hline
    YOLOv5s6 &797 ms& 156 ms&71 ms&24 ms\\
    \hline
    YOLOv5n6 &388 ms&58 ms&53 ms&22 ms\\
    \hline
\end{tabular}
    \caption{Inference with Three YOLO Models on Different Computers: This evaluation illustrates a key advantage of YOLO: the ability to adapt to available computing resources by choosing the appropriate model for training and inference.\newline
    The model highlighted in \textit{italic} on the first line is the one used in this work.}
    \label{tab:training_yolo_inference}
\end{table*}

In \ref{app:frameworks}, we evaluated several computer vision frameworks and found YOLO to be particularly promising. Consequently, we conducted an in-depth analysis of YOLO's performance, focusing on three high-resolution models with varying complexities: YOLOv5l6, YOLOv5s6, and YOLOv5n6. We compared their training and inference times per image across four different computing environments: our GPU server with an NVIDIA RTX A5000 (used for both training and inference), a Raspberry Pi 4 with 4 GB of RAM, a standard computer with an Intel i7 processor and 12 GB of RAM, and a high-performance server with 36 Intel Xeon cores and 256 GB of RAM. Training was configured with a maximum of 400 epochs and included early stopping if no further improvement was observed.

We list the model complexity in Table~\ref{tab:training_yolo_complexity}, the training times in Table~\ref{tab:training_yolo_training} and the inference times in Table~\ref{tab:training_yolo_inference}. We marked the model we used in our case in italic. The results in those tables demonstrate how model complexity influences inference times across different platforms. For embedded systems, the nano models (e.g., YOLOv5n6) are a suitable choice due to their lower resource requirements, while more complex models can be utilized on more powerful machines.

Following training, we assessed the performance of the three YOLOv5 models based on their \fone-score. Table~\ref{tab:eval_yolo} shows that, similar to the training results, the most complex model (YOLOv5l6) outperformed the simpler versions.

YOLO offers three main advantages: it delivers high-quality object detection with minimal training effort, scales effectively across various computing resources, and benefits from active development with regular updates. Therefore, we decided to use YOLO in this work.

\begin{table}
    \centering
\begin{tabular}{ r r r r r r r }
    \textbf{Model} & \textbf{FP} & \textbf{TP} & \textbf{FN} & \textbf{Prec.} & \textbf{Recall} & \(\textbf{F}_{\textbf{1}}\)\\
    \hline
    \textit{yolov5l6} & \textit{97} & \textit{867} & \textit{622} & \textit{0.899} & \textit{0.582} & \textit{0.707} \\
    \hline
    yolov5s6 & 179 & 873 & 616 & 0.830 & 0.586 & 0.687 \\
    \hline
    yolov5n6 & 130 & 826 & 663 & 0.864 & 0.555 & 0.676 \\
    \hline
\end{tabular}
    \caption{Evaluation of YOLO Models on the Reference Dataset: This analysis demonstrates the impact of model size on performance, with more complex models generally yielding better results.\newline
        The model highlighted in \textit{italic} on the first line is the one used in this work.}
    \label{tab:eval_yolo}
\end{table}

\bibliographystyle{elsarticle-num}
\bibliography{misc.bib}

\end{document}